\documentclass[11pt,a4paper]{article}
\usepackage[hyperref]{emnlp2021}
\usepackage{times}
\usepackage{latexsym}
\usepackage{graphicx}
\usepackage{amsmath}
\usepackage{amsfonts}
\usepackage{multirow}
\usepackage{bm}
\usepackage[english]{babel}
\usepackage{booktabs}
\usepackage{url}
\usepackage{array}

\usepackage{caption}
\usepackage{booktabs}
\usepackage{bbding}
\usepackage{pifont}
\usepackage{wasysym}
\usepackage{amssymb}
\usepackage{color}
\usepackage{colortbl}
\usepackage{algorithm}
\usepackage{algorithmicx}
\usepackage{algpseudocode}

\usepackage{soul}

\definecolor{mygray}{gray}{.9}

\newcommand{\PreserveBackslash}[1]{\let\temp=\\#1\let\\=\temp}
\newcolumntype{C}[1]{>{\PreserveBackslash\centering}p{#1}}
\newcolumntype{R}[1]{>{\PreserveBackslash\raggedleft}p{#1}}
\newcolumntype{L}[1]{>{\PreserveBackslash\raggedright}p{#1}}

\title{e-CARE: a New Dataset for Exploring Explainable Causal Reasoning}
\author {\textbf{Li Du, Xiao Ding\thanks{Corresponding author}, Kai Xiong, Ting Liu, and Bing Qin} \\
        Research Center for Social Computing and Information Retrieval \\
        Harbin Institute of Technology, China \\
        \{ldu, xding, kxiong, tliu,qinb\}@ir.hit.edu.cn
        }
        
\date{}

\begin{document}
%\linenumbers
\maketitle
\begin{abstract}
Understanding causality has vital importance for various Natural Language Processing (NLP) applications. Beyond the labeled instances, conceptual explanations of the causality can provide deep understanding of the causal facts to facilitate the causal reasoning process. However, such explanation information still remains absent in existing causal reasoning resources. In this paper, we fill this gap by presenting a human-annotated explainable CAusal REasoning dataset (e-CARE), which contains over 21K causal reasoning questions, together with natural language formed explanations of the causal questions. 
%We present \textit{e}CARE, an human-annotated eXplainable Causal Reasoning dataset, which contains over 18K causal reasoning questions, together with natural language formed conceptual explanations of the causal questions. 
Experimental results show that generating valid explanations for causal facts still remains especially challenging for the state-of-the-art models, and the explanation information can be helpful for promoting the accuracy and stability of causal reasoning models. 
%Our work highlights the importance of conceptual explanation in causal reasoning, and leads to new insights about more powerful and stable causal reasoning systems.

\end{abstract}

\newcommand{\tabincell}[2]{\begin{tabular}{@{}#1@{}}#2\end{tabular}}

%% big picture!!
%%

\section{Introduction}

%Causal reasoning is the process of capturing and understanding the causal dependencies amongst event and actions.

%Causal reasoning is one of the most central cognitive ability of human beings \cite{waldmann2013causal,jonassen2008designing}, which enables one to understand the observed facts and predict the future. Hence, understanding the causality and making causal predictions can be critical important for various NLP applications, such as decision making \cite{zhang2018fairness}, dialogue generation \cite{oh2013question}, and question answering \cite{oh2013question}.

%Human beings are remarkably good at learning and inferring the commonsense causality, as Psychological and  physiological experiments show that, in daily life people can reliably induce causality from just a few observations, and generalize it to unseen cases to infer the causation \cite{read1983once,griffiths2009theory}. 
Causal reasoning is one of the most central cognitive abilities of human beings \cite{waldmann2013causal,jonassen2008designing}, which enables one to understand the observed facts and predict the future. However, although recent causal reasoning models have achieved impressive performances on certain hand-crafted datasets, there still remains a considerable gap compared to human performances, as they cannot achieve stable performances across different datasets and are susceptible to adversarial attacks \cite{mccoy2019right,poliak2018collecting,gururangan2018annotation}. 

\begin{figure}
    \centering
    \includegraphics[width=0.99\linewidth]{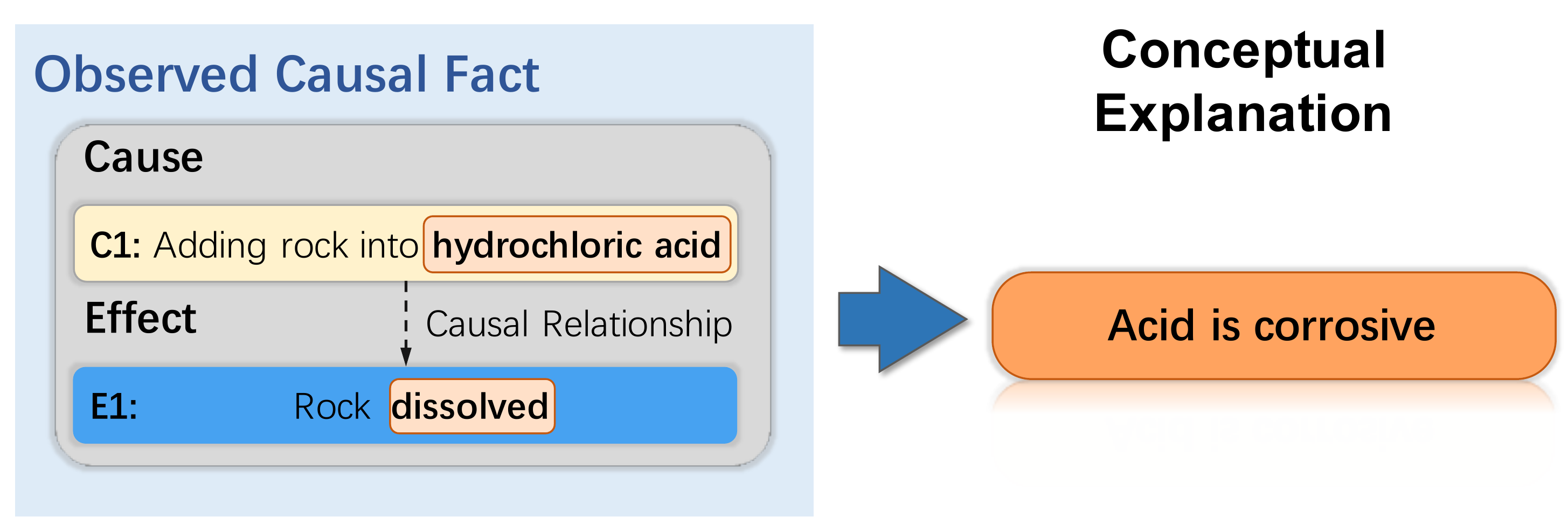}
    %\vspace{-0.5cm}
    \caption{Conceptual explanations of observed causality can be helpful for understanding the unseen causal facts.}
    \label{fig:example}
    %\vspace{-0.5cm}
\end{figure}

One key factor leading to such drastic contrast is that, present causal reasoning models only learn to induce empirical causal patterns that are predictive to the label, while human beings seek for deep and conceptual understanding of the causality to explain the observed causal facts.  
%beyond the specific concrete causality.
%The conceptual explanations can not only enables explicit justification for the understanding of the causal fact, but can And the correct conceptual explanations can in turn support the causal reasoning process. 
The conceptual explanations can not only serve as a touchstone to examine whether the underlying causal mechanism has been thoroughly understood, but it can also in turn support the causal reasoning process.
As illustrated in Figure~\ref{fig:example}, observing the causal fact $C_1$: \emph{adding rock into hydrochloric acid} causes $E_1$: \emph{rock dissolved}, one may further ask \emph{why such a causal relationship exists} and reach the plausible \emph{conceptual explanation} that \emph{Acid is corrosive}, which goes beyond the isolated facts and reaches the conceptual nature to reveal the principle of the causal mechanism. %Hence, through reaching for the conceptual explanation, on the one hand, it promotes one to have deep understanding of the causality, which is helpful for future causal reasoning; on the other hand, it enables an explicit justification of whether the causality is correctly and thoroughly understood. 
%The conceptual explanation reveals the principle of the causal mechanism through describing the corrosive nature of acid. 

However, despite the critical importance of conceptual explanations in causal reasoning, there is still a lack of such an explainable causal reasoning dataset. 
%The absent of such conceptual explanations would restrict the training and justification of causal reasoning systems that have deep understanding of causal mechanism to conduct explanative causal reasoning. 
To fill this gap, we contribute an explainable CAusal REasoning dataset (e-CARE),together with a new causal explanation generation task, and a novel Causal Explanation Quality (CEQ) evaluation metric.  
%built through crowdsourcing, together with a new causal explanation generation task, and a novel automatic score CEQ (Causal Explanation Quality) for evaluating the quality of generated explanations. 

The e-CARE dataset is constructed by crowdsourcing and contains over 21K multiple-choice causal reasoning questions, which makes e-CARE the largest human-annotated commonsense causal reasoning dataset to the best of our knowledge. In addition to the causal reasoning question itself, e-CARE also provides a free-text-formed conceptual explanation for each causal question to explain why the causation exists. On this basis, we propose a new causal explanation generation task that requires models not only to choose the correct causal fact but also to generate the explanation for the choice.
%As Table~\ref{tab:exp} shows, given a premise, the causal reasoning question requires choosing a plausible hypothesis from two candidates, so that the premise and hypothesis can form into a valid causal fact. 
%In addition, the causal fact is accompanied by a free-text-formed conceptual explanation about why the causal fact can exist. 
%Based on the additional explanations, we further propose a causal explanation generation task to examine model's understanding about causal mechanism. 
In addition, to directly measure the quality of generated explanations, we propose a novel causal explanation quality evaluation metric (namely, CEQ score). Compared to conventional text generation evaluation metrics such as BLEU \cite{papineni2002bleu} and ROUGE \cite{lin2004rouge} which mainly evaluate the textual or semantic similarity between generated explanations with golden annotations, CEQ score focuses on evaluating how much promotion an explanation can bring to understanding the causal mechanism. The dataset is publicly available at \url{https://github.com/Waste-Wood/e-CARE/}.

%We then investigate the efficacy of the conceptual explanations in causal reasoning, and the viability of generating the conceptual explanations.  
%indicates the effectiveness of the e-CARE dataset in evaluating the un of model.
Experimental results demonstrate that the causal questions of e-CARE are still challenging for the state-of-the-art (SOTA) pretrained language models, indicating the effectiveness of the e-CARE dataset in evaluating the causal learning ability of models.
In addition, the explanation signal received in the training process can enhance the performance and the stability of the reasoning model, while the SOTA baselines still have trouble in explaining the causal facts at a conceptual level.
%Experimental results show that the explanation signal received in the training process can enhance the performance and the stability of the reasoning model. While the state-of-the-art models still have trouble to explain the causal facts at a conceptual level. 
These analyses highlight the importance of the conceptual explanations in causal reasoning, and suggest an avenue for future researches.

%the causal reasoning task can still be challenging for models still have trouble to explain the causal reasoning. why an answer is correct,

\iffalse
\begin{table*}[t]
    \centering
    \scriptsize
    \begin{tabular}{l|l}
    \toprule
    \textbf{(a)} \textit{Premise}: Tom holds a copper block by hand and heats it on fire. & \textbf{(b)} \textit{Premise}:This computer's heat dispersion performance is bad. \\
    \quad \ \ \textit{Ask-for}: Effect & \quad \ \ \textit{Ask-for}: Effect \\
    \quad \ \ \textit{Hypothesis 1}: His fingers feel burnt immediately. (\Checkmark) & \quad \ \ \textit{Hypothesis 1}: Designers add copper tube into the computer. (\Checkmark) \\
    \quad \ \ \textit{Hypothesis 2}: The copper block keeps the same. (${\times}$) & \quad \ \ \textit{Hypothesis 2}: Designers put the computer into the ice water. \ \ (${\times}$) \\
    \hline
    \quad \ \ \textit{Explanation}: \textbf{Copper is a good thermal conductor.} & \quad \ \ \textit{Explanation}: \textbf{Copper is a good thermal conductor.} \\ 
    \bottomrule
    \end{tabular}
    \caption{Two instances from the e-CARE dataset. }
    \label{tab:exp}
%\vspace{-0.1cm}
\end{table*}
\fi

\section{Related Work}
\subsection{Commonsense Causal Reasoning Datasets}

%Due to the vital importance of causality knowledge, previous work have been dedicated to provide evaluations of commonsense causal reasoning ability. 
Existing commonsense causal reasoning corpora differ in their annotation guidelines and how they are constructed: (1) whether the corpus is automatically constructed or built by human annotation; (2) whether the annotation unit of the corpus is word-level, phrase-level, or sentence-level.

To obtain abundant causal knowledge, a natural way is extracting causal knowledge using heuristic rules from large-scale open-domain web text corpora \cite{luo2016commonsense,li2020guided,sap2019atomic}. 
%For example, \citet{luo2016commonsense} induce a word-level causality knowledge base CausalNet using hand-crafted rules from large-scale web text. While \citet{li2020guided} propose CausalBank, which extends the scale of CausalNet by adding phrase-level causal knowledge. However, the automated constructed dataset may subject to the significant reporting bias brought by the prior settings of building process \cite{sap2019atomic}. 
However, the reporting bias may challenge both the coverage and quality of the extracted causal knowledge.

Different from automatic construction, human annotation can endow datasets with higher precision. A line of work focuses on providing word-level causality knowledge \cite{girju2007semeval,mostafazadeh2016caters,do2011minimally,hendrickx2019semeval}. However, a word is not a complete semantic unit, which may limit the integrity of causal expressions and lead to ambiguity. To address this issue, other datasets are constructed to provide phrase-level \cite{caselli2017event,bethard2008learning,mirza2014annotating,dunietz2017because} and sentence-level \cite{ning2019joint,roemmele2011choice} causal knowledge. Among these datasets, COPA \cite{roemmele2011choice} has become a widely adopted benchmark. Nevertheless, the size of COPA is rather limited, which may result in over-fitting and arouse concerns about the confidence of the results. 
%Table 5 contrasts the size of causal portion of prior resources with our own.

In this paper, we introduce an \textit{e}xplainable CAusal REasoning dataset (e-CARE). As shown in Table~\ref{tab:pw}, to the best of our knowledge, e-CARE is the largest human-annotated causal reasoning dataset. With more than 21,000 instances, the e-CARE dataset can serve as a more reliable benchmark. Furthermore, compared to previous work, e-CARE can provide additional explanation information, which plays a critical role in learning the underlying mechanism of causal knowledge.

\begin{table}[t]
    \centering
    \small
    %\begin{tabular}{ll}
    \begin{tabular}{p{4.68cm}p{0.35cm}p{0.6cm}p{0.35cm}}
    \hline
    \multicolumn{4}{l}{\textbf{Dataset} \qquad \qquad \qquad \qquad \qquad \ \textbf{Anno. Unit} \ \textbf{Size}  \ \textbf{Expl.} } \\  
    \hline
    \rowcolor{mygray} \textit{Automatically-Built Dataset} & & &\\
    %\hline
    CausalNet \cite{luo2016commonsense} & W & 11M & N \\
    CausalBank \cite{li2020guided} & P & 314M & N \\
    \hline
    %\multicolumn{3}{l}{\textbf{Human-Annotated Dataset}}  \\
    \rowcolor{mygray} \textit{Human-Annotated Dataset} & & & \\
    SemEval-2007 T4 \scriptsize{\cite{girju2007semeval}} & W & 220 & N \\
    CaTeRS \scriptsize{\cite{mostafazadeh2016caters}} & W & 488 & N \\
    EventCausalityData \scriptsize{\cite{do2011minimally}} & W & 580 & N \\
    SemEval-2010 T8 \scriptsize{\cite{hendrickx2019semeval}} & W & 1,003 & N \\
    ESC \scriptsize{\cite{caselli2017event}} & P & 117 & N \\ 
    T-CBank \scriptsize{\cite{bethard2008learning}} & P & 271 & N \\
    CausalTimeBank \scriptsize{\cite{mirza2014annotating}} & P & 318 & N \\
    BECauSE 2.0 \scriptsize{\cite{dunietz2017because}} & P & 1,803 & N \\
    TCR \scriptsize{\cite{ning2019joint}} & S & 172 & N \\
    COPA \scriptsize{\cite{roemmele2011choice}} & S & 1,000 & N \\
    %\hline
    %\rowcolor{mygray} 
    %\multicolumn{3}{l}{\textbf{Automatically-Built Dataset}} \\
    \hline
    \textbf{e-CARE} & \textbf{S} & \textbf{21K} & \textbf{Y} \\
    \hline
    \end{tabular}
    \caption{A list of previous commonsense causal reasoning datasets. In the column ``Annotation Unit'', ``W'', ``P'' and ``S'' are abbreviation of word, phrase and sentence, respectively. ``Expl.'' is the abbreviation of ``Explanation''.
    }
    \label{tab:pw}
%\vspace{-0.1cm}
\end{table}

%Our new resource ART complements on going efforts in building resources for natural language inference

\subsection{Explainable Textual Inference}

Recently, an increasing amount of datasets have been proposed to address the explainability of textual inference tasks, such as textual entailment inference \cite{camburu2018snli}, question-answering (QA) \cite{deyoung2019eraser,perez2019finding} and multi-hop QA \cite{ye2020teaching}. The form and content of the explanations vary with the nature of specific tasks.

The QA task requires a model to answer the question based on evidences within given texts. Therefore, the explanation for this task should describe where and how an answer can be found \cite{wiegreffe2021teach}. The explanations can have various forms, including answer-bearing sentences \cite{perez2019finding}, structured information connecting the question and answer \cite{hancock2018training,ye2020teaching}, or even human-annotated free-formed sentences \cite{camburu2018snli,rajani2019explain}.
In contrast, the multi-hop QA task requires the model to infer the correct answer through multiple reasoning steps. Hence, the explanation of this task needs to provide the specific reasoning paths  \cite{wiegreffe2021teach,jhamtani2020learning}. %For example, 
%For tasks requiring multi-hop inference process, an explanation is often taken as the chain of reasoning steps leading to an answer. 
%\citet{inoue1910rc} point out the answer-supporting sentences in the given text as the explanation, 
%\citet{khot2020qasc} provide an human-annotated reasoning chain as an explanation for each multi-hop QA question. While \citet{jhamtani2020learning} argue that one multi-hop QA question may corresponds to multiple plausible reasoning chains. So they obtain a collection of reasoning chains by human annotation to construct an explanation set for each question. 
%, but there are two critical distinctions that make abductive reasoning uniquely challenging. First,

Our work is quite different from previous work. We notice that all of these previous work only offer explanations that explain a specific question. Whereas we aim at providing a conceptual understanding of the causality, which has the potential to explain \emph{a set of related causal observations}, rather than only explain a specific causal fact. 
%With the increased generality, the conceptual explanations can offer stronger support for the reasoning process. 
%On the other hand, the conceptual explanation generation task can be a critical test for model's ability in obtaining deep understandings of the observed causal facts.

\begin{table}[t]
    \centering
    \small
    \begin{tabular}{lcccc}
    \hline
    %\toprule
    \textbf{Number}     & \textbf{Train} & \textbf{ Dev} & \textbf{Test} & \textbf{Total}\\
    \hline
    %\midrule
    %\rowcolor{mygray} 
    %\textit{Total unique occurrences} &  &  & \\
    Causal Questions & 14,928 & 2,132 & 4,264 & 21,324 \\    
    Uniq. Explanations & 10,491 & 2,102 & 3,814 & 13,048 \\
    \hline
    %\bottomrule
    \end{tabular}
    \caption{Corpus level statistics of the e-CARE dataset. Uniq. Explanations refer to the explanations that only correspond to a single causal fact.}
    \label{tab:stat}
%\vspace{-0.1cm}
\end{table}

\section{e-CARE: an Explainable Causal Reasoning Dataset}
%is the first large-scale dataset investigating the explainability of causal reasoning. It
e-CARE contains a total of 21,324 instances, corresponding to 13,048 unique explanations. This also makes e-CARE the largest human-annotated commonsense causal reasoning benchmark. The corpus-level statistics of the e-CARE dataset are shown in Table~\ref{tab:stat}. 

As shown in Table~\ref{tab:exp}, each instance of the e-CARE dataset is constituted by two components: (1) a multiple-choice causal reasoning question, composed of a premise and two hypotheses, and one of the hypotheses can form a valid causal fact with the premise; (2) a conceptual explanation about the essential condition that enables the existence of the causal fact. For example, as Table~\ref{tab:exp} shows, the \emph{explanation} points out the nature of copper that \emph{Copper is a good thermal conductor}, so that holding copper on fire will make fingers feel burnt immediately. The appendix provides more discussion about the explanations within e-CARE. On this basis, we introduce two tasks:

%Note that, in this paper, we do not assume the explanation to be self-sufficient. In other words, 

\noindent \textbf{Causal Reasoning Task}
We formulate the causal reasoning task as a multiple-choice task: given a premise event, one needs to choose a more plausible hypothesis from two candidates, so that the premise and the correct hypothesis can form into a valid causal fact.

\noindent \textbf{Explanation Generation Task}
It requires the model to generate a free-text-formed explanation for a given causal fact (composed of a premise and the corresponding \emph{correct} hypothesis).

%While as illustrated in Table~\ref{tab:exp}~(b), the same explanation can also provide insights about another causal fact seemingly totally different from the case in Table~\ref{tab:exp}~(a), that putting copper tubes into computer can promote thermal dispersion. This demonstrate the usefulness of the conceptual explanations in providing the deep understanding of causality to support the causal reasoning.

\subsection{Data Annotation}
%Rather than first build the causal questions then generate the corresponding explanations, the e-CARE dataset is built in a reverse manner. 
%In specific, we first acquire some statements, describing certain . Then for each statement, we collect a set of causal-effect pairs which can be explained by the statement, and build the causal reasoning questions based on the causal causal-effect pairs.
%construct the e-CARE dataset in a top-down manner,
To construct the e-CARE dataset, we start by collecting statements that describe conceptual understandings of world knowledge. Then given a statement, we ask different annotators to generate causal facts that can be explained by the statement, and build causal questions based on these causal facts. 
This is because we hope to provide conceptual explanations with more generality, that can explain a set of correlated causal facts, instead of only applicable to a certain isolated causal fact.
%While finding the causal questions that can be explained by a same explanation from a large causal reasoning question set is almost impractical, 
%Hence, if the causal fact set was collected at first, then it would be almost impractical to find out the causal facts that share a same explanation from the large-scale causal fact set, as this searching process has an $\mathcal{O}(N^2)$ time complexity. On the contrary, 
Moreover, the statements can serve as clues to help the annotators to come up with causal facts.

\begin{table}[t]
    \centering
    \small
    \begin{tabular}{l}
    %\toprule
    \hline
    \textit{Premise}: Tom holds a copper block by hand and \\ \quad \quad \quad \quad  heats it on fire.  \\
    \textit{Ask-for}: Effect \\
    \textit{Hypothesis 1}: His fingers feel burnt immediately. (\Checkmark)  \\
    \textit{Hypothesis 2}: The copper block keeps the same. \ (${\times}$) \\
    \hline
    \textit{Explanation}: \textbf{Copper is a good thermal conductor.}  \\ 
    %\bottomrule
    \hline
    \end{tabular}
    \setlength{\belowcaptionskip}{-0.3cm}
    \caption{An instance from the e-CARE dataset. }
    \label{tab:exp}
%\vspace{-0.1cm}
\end{table}

%it can be a natural process for a human annotator to come up with some causal facts that can be explained by a statement about certain world knowledge.

%while different people may have different understanding to the same causality relationship. And the same explanation can be expressed in multiple ways. Which makes that, if it will be pretty hard to find  

%Therefore, we choose to , so that we can obtain a set of relevant causal questions that can be explained by the same generic statements.
%Then human annotators are the causal that can be explained by the generic statement.

%guideline !!!

\noindent \textbf{Collecting Potential Explanations}
Two key issues remain in collecting statements as potential explanations: (1) what kind of statements can be potential conceptual explanations of the causal facts; (2) where to find the appropriate statements. 

%in general, to be explanative to causal facts, a statement provides three kinds of information:

For the first question, \citet{jonassen2008designing} concluded that, in general, the explanation of causality mainly describes three categories of information: (1) the nature or attributes of the objectives involved in the causal facts; (2) forces or actions that cause changes and drive transient motions; (3) the goals, intentions, motives or purposes of the causal agents. In addition, to be the conceptual explanation of a causal fact, the statement should be able to involve with a category of objects or people, but not only focus on a specific object or person \cite{sembugamoorthy1986functional}.
%Hence, to be the explanation of causal facts, the statement must describe some \emph{general truth}, that 
%which distinguishes with the statements that describes 
%, or the statements that holds true only under certain context.

%The above-mentioned two principles suggest what kind of statements can potentially be qualified explanations for causal facts. 
Following these principles, we notice that there are already several available knowledge bases containing statements about such generic world knowledge, including ConceptNet \cite{speer2013conceptnet}, WordNet \cite{fellbaum2010wordnet}, Atomic \cite{sap2019atomic} and GenericsKB \cite{bhakthavatsalam2020genericskb}. However, ConceptNet and WordNet are structured knowledge graphs, containing only triplet-structured statements with a limited number of predicates. The scope of Atomic is limited in the activities of human beings. Compared to these knowledge bases, GenericsKB is an open-domain, large-scale knowledge base, containing rich generic world knowledge described in free-form text. Therefore, we collect the statements from GenericsKB to ensure the coverage and diversity of the potential explanations.
%Instead of , we propose to collect the statements from the already existed knowledge bases   

Specifically, we filter out the statements in GenericsKB with low reliability, and the statements that may disobey the above-mentioned three principles. More details are provided in the Appendix. Thereafter, a total of 19,746 statements are left to form into a potential explanation set, which is further provided to the annotators to generate the causal questions. 

\noindent \textbf{Annotating Causal Reasoning Questions}
Given the potential explanation set, annotators were recruited to generate corresponding causal questions. Specifically, a causal question is generated by two steps: 

First, an annotator was presented with a statement as a potential explanation, and was instructed to write a causal fact (composed of a cause and an effect), so that the causal fact can be interpreted by the given statement. In this step, a key issue is controlling the quality of generated causal facts. Thus we demonstrated illustrative examples to guide the annotators to avoid the following mistakes: 

(1) The created cause and effect are not in a valid causal relationship; 

(2) The created causal fact cannot be explained by the provided statement; 

%(3) The statement describes a middle step of the causal reasoning process;

(3) There are factual errors or imaginary contents in the created causal facts.

In the causal fact generation process, each statement is randomly distributed to 1-3 annotators, so that we can find some statements that could explain multiple causal facts. Note that, in this process, we do not assume all statements are necessary to be a valid explanation. In other words, we do not require that the annotators must generate a causal fact for each given statement. Instead, we leave it to the judgment of annotators. In this way, the unreliable statements can be further excluded to promote the quality of our dataset. 

After the generation of causal facts, an ask-for indicator $a\in$ [``cause'', ``effect''] was randomly generated, where $a=$ ``cause'' (``effect'') means that the cause (effect) event is the hypothesis, and the effect (cause) event is the premise of the causal question, respectively. Then given the ask-for indicator, in order to control the grammar and writing style consistency, the same annotator was prompted to write a distract cause (effect) as the implausible hypothesis according to the ask-for indicator. In this process, the annotators were instructed to create the implausible hypothesis as close as  possible to the true hypothesis, meanwhile prevent creating uninformative distractors (such as simply adding a ``not'' into the true hypothesis). 
%Finally, we also asked workers to verify if the
%More details about the collection of causal reasoning questions can be found in the appendix.

\begin{table}[t]
    \centering
    \small
    \begin{tabular}{lcc}
    \hline
    \textbf{Model} & \textbf{Dev} & \textbf{Test}\\
    \hline
    Random & 50.1 & 50.1 \\
    GPT2 \cite{radford2018improving}& 57.17  & 56.30  \\
    RoBERTa \cite{liu2019roberta} & 58.38 & 56.42 \\
    BERT \cite{devlin2019bert} & 56.19 & 54.45 \\ 
    \hline
    \end{tabular}
    \setlength{\belowcaptionskip}{-0.31cm}
    \caption{Model's accuracy (\%) of choosing the correct hypothesis without the premise.}
    \label{tab:h-}
%\vspace{-0.1cm}
\end{table}

\subsection{Refinement and Analysis of the e-CARE Dataset}

%A significant challenge in dataset construction is avoiding \emph{annotation artifacts}, i.e., the unintentional features within the dataset that leak information about the target label \cite{gururangan2018annotation,poliak2018collecting}. To address this issue, following \citet{bhagavatula2019abductive} and \citet{sakaguchi2020winogrande}, for each causal question, we further collected multiple distractor hypotheses, and employed an adversarial filtering algorithm to retain the most indistinguishable distractor. More details about the adversarial filtering algorithm are provided in the appendix. The corpus level statistics of e-CARE dataset after adversarial filtering are shown in Table~\ref{tab:stat}. Comparison between the number of total explanations and the number of explanations that only correspond to a single causal fact (denoted as Uniq. Explanations) show that, among the e-CARE dataset, most explanations correspond to at least two causal facts.

A significant challenge in dataset construction is avoiding introducing \emph{superficial cues} into the dataset \cite{gururangan2018annotation,poliak2018collecting}, which refers to the unintentional features that leak the label information. To address this issue, following \citet{bhagavatula2019abductive} and \citet{sakaguchi2020winogrande}, we employ an adversarial filtering algorithm to replace the implausible hypotheses that can easily be distinguished with the correct hypotheses using the superficial clues.
More details about the adversarial filtering are provided in the Appendix. As Table~\ref{tab:h-} shows, after the adversarial filtering, without the existence of the premise, the SOTA pretrained language models can hardly distinguish two candidate hypotheses, which indicates that to predict the correct label, a model must understand the causal relationship between the premise and hypothesis, rather than only depend on the superficial cues within the two hypotheses.
%Comparison between the number of total explanations and the number of explanations that only correspond to a single causal fact (denoted as Uniq. Explanations) show that, among the e-CARE dataset, most explanations correspond to at least two causal facts.

After the refinement, we evaluate the quality of the annotated causal questions and collected explanations through crowdsourcing. We assess the quality of causal questions by testing if there is agreement among human raters on the answer of causal questions. Specifically, we randomly sampled 200 causal questions from e-CARE, and enlisted 10 annotators to answer the causal questions. In this process, each causal question was evaluated by three annotators. When answering the causal questions, the raters were allowed to choose an additional option ``None of the above'' if neither hypothesis was deemed plausible. The human annotators achieve a 92\% accuracy with a high agreement (Cohen's $\kappa$ = 0.935) \cite{cohen1960coefficient}.

To validate the quality of explanations, we enlisted volunteers to determine whether or not the explanations can explain corresponding causal facts. In total 200 causal facts with corresponding explanations were sampled and distributed to 10 volunteers, and each explanation was evaluated by three volunteers. After the evaluation, on average 89.5\% of the explanations were deemed as valid (Cohen's $\kappa$ = 0.832), showcasing the quality of the explanations in e-CARE.

\section{Causal Explanation Quality (CEQ) Score} 

A number of automatic scores have been proposed to evaluate the quality of generated explanations, such as BLEU \cite{papineni2002bleu} and ROUGE \cite{lin2004rouge}. However, these metrics evaluate the quality of the generated explanations only through comparing the textual or semantic similarity between the generated explanations and the golden annotation. Alternatively, an ideal causal explanation quality evaluation metric should directly measure if the causal fact is appropriately explained by the explanation. 

Hence, we propose a novel causal explanation quality evaluation metric (namely, CEQ score) as a step towards directly measuring the quality of generated explanations. We devise the CEQ score based on the consideration that a better explanation should provide more information for understanding the causality, so that the prediction model can more accurately estimate the reasonableness of the causal fact. Previous literature characterized such reasonableness as the \emph{causal strength} of the given causal fact \cite{roemmele2011choice,luo2016commonsense}, where the \emph{causal strength} is a score in $[0,1]$. Hence, in theory, for a valid causal fact, its causal strength should be equal to 1. Given a valid causal fact, an explanation should help to increase its estimated causal strength to the ground-truth value 1.

%that a better explanation should provide more information for understanding the causality, so that the prediction model can more accurately estimate the \emph{causal strength} of given causal fact, where the \emph{causal strength} belongs to $[0,1]$, measuring the probability that a causal fact is valid. For a valid causal fact, ideally, its causal strength should equal to 1. Hence given a valid causal fact, an explanation should help to increase the estimated causal strength to the ground truth value 1.

Therefore, we can evaluate the quality of a generated explanation by measuring the increase of causal strength brought by the explanation. 
Specifically, let $C$, $E$, and $X$ denote the cause, the effect and the generated explanation, respectively. Formally, the CEQ score is defined as:
\begin{equation}
%\small
\mathrm{CEQ}=\Delta_{\mathrm{cs}}=\mathrm{cs}(C,E|X)-\mathrm{cs}(C,E),
\end{equation}
where $\mathrm{cs}(C,E)$ is the original causal strength between $C$ and $E$; $\mathrm{cs}(C,E|X)$ is the causal strength after involvement of the additional explanation information. 
The explanation enhanced causal strength $\mathrm{cs}(C,E|X)$ is defined as: 
\begin{equation}
\small
\mathrm{cs}(C,E|X)=max[\mathrm{cs}(C+X,E), \mathrm{cs}(C, E+X)],
\end{equation}
where ``+'' denotes the string concatenate operation. Therefore, the CEQ score is positively related to the \emph{increase of causal strength} between $C$ and $E$ after the involvement of the explanation $X$.

In this paper, we employ a widely-adopted model-agnostic method proposed by \citet{luo2016commonsense} to calculate the causal strength. The model-agnostic nature enable us to avoid reliance on certain models and keep the fairness of evaluation. Specifically, the phrase-level causal strength is derived through synthesizing the word-level causality. 
\begin{equation}
\small
\it \mathrm{cs}(C_A,E_B)=\frac{1}{N_{C_A} + N_{E_B}}{\sum_{w_i\in C_A, w_j\in E_B}\mathrm{cs}(w_i, w_j)},
\end{equation}
where $\it (C_A,E_B)$ is an arbitrary causal fact; $N_{C_A}$ and $N_{E_B}$ are the number of words within $C_A$ and $E_B$, respectively; $\mathrm{cs}(w_i, w_j)$ is the causal strength between word $w_i$ and $w_j$, which is estimated from a large corpus as: 
\begin{equation}
\small
%\mathrm{cs}(w_i, w_j)=\frac{\mathrm{Count}(w_i, w_j) / M}{\mathrm{Count}(w_i)/N\mathrm{Count}(w_j)^{\alpha}/N},
\mathrm{cs}(w_i, w_j)=\frac{\mathrm{Count}(w_i, w_j)}{\mathrm{Count}(w_i)\mathrm{Count}(w_j)^{\alpha}},
\end{equation}
where $\alpha$ is a penalty coefficient and \citet{luo2016commonsense} empirically set $\alpha=0.66$.
%, $M$ is the total number of word pairs within the corpus, $N$ is the vocabulary size of the corpus.

\section{Experiments and Results}

We examine the performance of state-of-the-art pretrained language models on the causal reasoning task and the explanation generation task. Furthermore, we investigate the specific role of explanations in causal reasoning by: (1) a predict-and-generate experiment, which requires models to conduct the causal reasoning task and generate corresponding explanations simultaneously; (2) a stability analysis using adversarial attacks.

\begin{table}[t]
    \centering
    \small
    \begin{tabular}{lc}
    \hline
    \textbf{Model} & \textbf{Accuracy (\%)} \\
    \hline
    %GPT \cite{radford2019language} & \\
    GPT2 \cite{radford2019language} & 69.51 \\
    RoBERTa \cite{liu2019roberta} & 70.73 \\
    BART \cite{lewis2020bart} & 71.65 \\
    XLNET \cite{yang2019xlnet} & 74.58 \\
    BERT \cite{devlin2019bert} & 75.38 \\
    ALBERT \cite{lan2019albert} & 74.60 \\
    \hline
    \textbf{Human Performance} & \textbf{92.00} \\
    \hline
    \end{tabular}
    \caption{Performance of pretrained language models on the test set of the causal reasoning task.}
    \label{tab:accu_ca}
%\vspace{-0.5cm}
\end{table}

\begin{table*}[ht]
    \centering
    \small
    \begin{tabular}{lccccc}
    \hline
    \textbf{Model} & \textbf{AVG-BLEU} & \textbf{ROUGE-l} & \textbf{PPL} & \textbf{CEQ} & \textbf{Human Evaluation (\%)} \\
    \hline
    GRU-Seq2Seq & 18.66  & 21.32 & 33.71 & 0.024 & 0 \\
    GPT2 \cite{radford2019language} & 32.04 & 31.47 & 7.14 & 0.105 & 20.0 \\
    \hline
    Human Generation & \textbf{35.51} & \textbf{33.46} & \textbf{-} & \textbf{0.144} & \textbf{89.5}\\
    \hline
    %\textbf{Human Performance} & \\
    %\hline
    \end{tabular}
    \caption{Model performance on the explanation generation task.}
    \label{tab:res_eg}
%\vspace{-0.1cm}
\end{table*}

\subsection{Causal Reasoning}

%We evaluate the performance of SOTA pretrained language models on the causal reasoning task of the e-CARE dataset. 
%As Table~\ref{tab:exp} shows, the causal reasoning task can be characterized as a multiple-choice task.

\noindent \textbf{Settings} 
%To choose a plausible hypothesis so that the premise and hypothesis can form into a valid causal fact,
We cast the causal reasoning task as a prediction problem: The input of the model is a candidate causal fact composed of a premise and one of the corresponding candidate hypotheses. The output is a score measuring the reasonableness of the candidate causal fact. 
%Pretrained language models have demonstrated promising performances on various causal reasoning datasets \cite{wang2018glue,wangsuperglue}. 
We evaluate the causal reasoning ability of several SOTA pretrained language models, including discriminative pretrained language models BERT \cite{devlin2019bert}, RoBERTa \cite{liu2019roberta}, XLNet \cite{yang2019xlnet}, and ALBERT \cite{lan2019albert}; as well as autoregressive generative pretrained language models GPT2 \cite{radford2019language} and BART \cite{lewis2020bart}, which can also be adapted to the predictive causal reasoning task. In this section and the following parts, all experiments are conducted using the base-sized version of the pretrained language models. Additional details about experimental settings are provided in the Appendix.
%We adapt these two models to causal reasoning task by adding an $<\text{EOS}>$ token to the end of input text, and making predictions based on the representation of the $<\text{EOS}>$ token. In this section and following parts, we conduct experiments using the base-sized language models.
%As Table~\ref{tab:accu_ca} shows, on causal reasoning task human beings achieve an accuracy of \%.

\noindent \textbf{Results} 
%Since in the e-CARE dataset, the causal reasoning task is formulated as a binary classification problem, we evaluate the model performance using prediction accuracy (\%) on the test set of e-CARE. 
As shown in Table~\ref{tab:accu_ca}, ALBERT achieves the highest accuracy of 73.86\% on the causal reasoning task of e-CARE. However, ALBERT can achieve an accuracy of 86.0\% on the widely adopted causal reasoning benchmark COPA by our implementation. This is mainly because, on one hand, previous causal reasoning datasets are too small to evaluate the genuine reasoning ability of the model. On the other hand, previous datasets may provide some superficial cues for the reasoning models to achieve superb performances. In contrast, e-CARE is the largest causal reasoning dataset that can provide enough test instances to evaluate the actual ability of the model. Moreover, in the annotating process of e-CARE, we introduced an adversarial filtering process to avoid the influence of superficial cues on the performances of reasoning models.
%This difference indicates that, the seemingly superb performances of the SOTA models on previous causal reasoning datasets may not be achieved by understanding of the causality, but by utilizing the exploitable superficial clues. It also indicates that, 
Hence, we believe that e-CARE dataset can serve as a new benchmark for effectively evaluating models' causal reasoning ability. We also notice that human beings can achieve an accuracy of 92.00\% on the e-CARE dataset. The large gap between the human performance and the pretrained language models suggests that the causal reasoning questions provided in our dataset still remain challenging, and calls for more powerful causal reasoning models.

\begin{table}
    \centering
    \small
    \begin{tabular}{lcc}
    \hline
    & \textbf{Corr. Coef with Human Eval.} & \textbf{P-value} \\ 
    \hline
    \textbf{AVG-BLEU} & 0.032 & 0.749 \ \ \\
     \textbf{ROUGE-l} & 0.021 & 0.836 \ \ \\
     \textbf{CEQ} & \textbf{0.247} & \textbf{0.013*} \\
    \hline
    %\textbf{Human Performance} & \\
    %\hline
    \end{tabular}
    \caption{Pearson Correlation coefficients between human evaluation and automatic scores. ``*'' denotes P-value $<$ 0.05.}
    \label{tab:res_coef}
%\vspace{-0.1cm}
\end{table}

\begin{table*}[t]
    \centering
    \scriptsize
    \begin{tabular}{lllc}
    \hline
    \textbf{Causal Facts} & \textbf{(Generated) Explanation} &\textbf{Human Annotation} &\textbf{CEQ} \\
    \hline
    \tabincell{l}{\textit{Cause:} He was infected with gram-positive bacteria. \\ \textit{Effect:} The doctor raised the lysozyme in his body.} & \tabincell{l}{Lysozyme is a chemical produced by the  \\ bacterium Bacillus thuringiensis to kill bacteria.} & \tabincell{l}{Lysozyme destroys cell wall of \\ bacteria.} & \tabincell{l}{0.193} \\
    \hline
    \tabincell{l}{\textit{Cause:} Pneumonia pervaded in this area. \\ \textit{Effect:} Many people died of Pneumonia.} & \tabincell{l}{Pneumonia is a leading infectious disease \sethlcolor{yellow}\hl{caused} \\ \sethlcolor{yellow}\hl{by a bacterium called Pseudomonas aeruginosa.}} & \tabincell{l}{Pneumonia is the most common \\ infectious disease leading to death.} & \tabincell{l}{0.069} \\
    \hline
    \tabincell{l}{\textit{Cause:} Tom wanted to prevent cancer. \\ \textit{Effect:} The doctor told him to eat more foods \\ containing Vitamin C.} & \tabincell{l}{\sethlcolor{pink}\hl{Vitamin C is a naturally occurring} \\ \sethlcolor{pink}\hl{amino acid found in many foods.}} & \tabincell{l}{Vitamin C slow the growth of small \\ cell lung cancer cells.} & \tabincell{l}{0.012} \\ \cline{2-4}
    \hline
    %\textbf{Human Performance} & \\
    %\hline
    \end{tabular}
    \caption{Examples of explanations generated by GPT2. We highlighted the factual mistakes within the generated explanations and the totally irrelevant explanation in yellow and pink, respectively.}
    \label{tab:exp_eg}
%\vspace{-0.1cm}
\end{table*}

\begin{table*}[t]
    \centering
    \small
    \begin{tabular}{l|c|cccc|c}
    %\hline
    %\textbf{Experiment} & \textbf{Causal Reasoning} & \textbf{Adversarial Attack } & \multicolumn{4}{c}{\textbf{Explanation Generation}} \\
    \hline
    \textbf{Model} & \textbf{Accu (\%)} & \textbf{AVG-BLEU} & \textbf{ROUGE-l} & \textbf{CEQ} & \textbf{Human Eval. (\%)}  & \textbf{$\Delta$Accu. (\%) after Adv. Attack} \\
    \hline
    GPT2$_{\text{CR}}$  & 69.51 & - & - & - & - & -6.40  \\
    GPT2$_{\text{EG}}$  & - & 32.04 & 31.47 & 0.035 & 20.0& -  \\
    GPT2$_{\text{CR-EG}}$  & \textbf{71.06}  & \textbf{34.83} & \textbf{34.22} & \textbf{0.042} & \textbf{26.5} & \textbf{-5.49} \\
    %\hline
    %GPT2$_{\text{CR w/ Golden E}}$  & \textbf{66.21}  & - & - & - & - & \textbf{66.21} \\
    %\hline
    %Human Annotation & \textbf{} & - & \textbf{} & \textbf{} & \textbf{0.051} & \textbf{86.2} \\
    %BART$_{\text{EG-CR}}$  & & & & & & \\
    \hline
    %\textbf{Human Performance} & \\
    %\hline
    \end{tabular}
    \caption{Model performance on the test set of Joint Causal Reasoning and Explanation Generation task.}
    \label{tab:res_mt}
%\vspace{-0.1cm}
\end{table*}

\subsection{Explanation Generation}

We investigate whether the model can generate correct explanations for given valid causal facts by training a GRU-based Seq2Seq model \cite{chung2014empirical}, and finetuning a generative pretrained language model GPT2 \cite{radford2019language} on the e-CARE dataset. Both models take the concatenation of the cause and effect as input. Please refer to the Appendix for more details.

\noindent \textbf{Evaluation Metrics} We automatically evaluate the quality of generated explanations using average-BLEU (n=4) \cite{papineni2002bleu}, ROUGE-l \cite{lin2004rouge}, Perplexity \cite{horgan1995complexity}, together with our proposed CEQ score.

\noindent \textbf{Human Evaluation} 
%Since automatic evaluation of generated language is an open research question \cite{liu2016not}, 
We also assess the quality of model-generated explanations through human evaluation. Specifically, we sampled 200 explanations generated by each method. Then three workers were shown with the generated explanations, together with corresponding causal facts, and were asked to label whether the generated explanation can explain the corresponding causal fact. 

%\noindent \textbf{Human Performance} We also evaluate the human performance on the causal reasoning task through crowdsourcing. Three hundreds of causal questions were randomly sampled from the test set of e-CARE. Then each causal question is given to three human annotators who were prompted to choose the more plausible hypothesis. 

\noindent \textbf{Quantitative Results} As shown in Table~\ref{tab:res_eg}, 89.5\% of human-written explanations are found to be valid, while the generative pretrained language model GPT2 only achieves a correctness of 20.0\%. 
The last row of Table~\ref{tab:res_eg} reports the score of held-out human-written explanations, which serves as a ceiling for model performance.
The significant gap indicates that, although GPT2 can achieve impressive performance on various natural language generation tasks, it still remains especially challenging for GPT2 to deeply understand the causal facts and then generate explanations like human beings. This may be one of the main obstacles hindering the further improvement of present causal reasoning models.

%In addition, we calculate the Pearson correlation coefficient $\rho$ between the human evaluation result and the CEQ score, and get $\rho=0.247, \text{P-value}=0.013$. The significant positive correlation indicates the rationality of the CEQ score.

Moreover, we measure the similarity between the automatic scores with the results of human evaluation using the Spearman correlation coefficient. As Table~\ref{tab:res_coef} shows, ROUGH-l and average-BLEU barely have a correlation with the results of human evaluation. This is because average-BLEU and ROUGH-l only implicitly evaluate the quality of generated explanations by measuring the textual similarity with the golden annotations. %Previous literature have also reported the deficiencies these automatic scores in evaluating the quality of generated texts in other domain \cite{liu2016not,lowe2017towards,zhang2019bertscore}.
Compared to average-BLEU and ROUGH-l, the CEQ score has a significant positive relationship with the human evaluation results. This indicates the efficiency of the CEQ score in evaluating the quality of generated explanations.
%that, by measuring how much promotion an explanation can bring on understanding the causal mechanism, the CEQ score can directly evaluate the quality of generated explanations in some extent.   

\noindent \textbf{Qualitative Analysis} In Table~\ref{tab:exp_eg}, we provide examples of explanations generated by GPT2. We observe that GPT2 can generate a reasonable explanation for some causal facts, while the generated explanations may still contain factual mistakes, or be totally irrelevant to the given causal fact (highlighted in yellow and pink, respectively). This indicate that the explanation generation still remains challenging for the GPT2 model.

\subsection{Joint Causal Reasoning and Explanation Generation}

To investigate the role of causal explanations in the causal reasoning process, we trained models to jointly conduct these two tasks.
%The causal reasoning and the explanation generation process are correlated, as to choose the correct hypothesis for a causal reasoning question, model should firstly have deep understanding of the causal mechanism, which serve as an explanation. Hence, we investigate the specific relationship between causal reasoning and explanation generation by training model to jointly conduct these two tasks.
%predict the label for a given causal question and then generate the corresponding explanation;
%We explore explanation generation in two settings: (a) predict the label for a given causal question and then generate the corresponding explanation; (b) a more natural way is first think of an explanation then predict the label of a causal question, as model should first understand the causal fact, and then the label can be naturally predicted. 

\noindent \textbf{Settings}
Since this task requires a model to predict a label meanwhile generate an explanation, we conduct the experiments using the GPT2 model, which can be adapted to conduct the predictive causal reasoning task and explanation generation simultaneously. We denote this multi-task finetuned GPT2 model as GPT2$_{\text{CR-GE}}$. Details for training GPT2$_{\text{CR-GE}}$ is provided in the Appendix.

To make the performance comparable, when evaluating the performance of GPT2$_{\text{CR-GE}}$ on the causal expatiations generation task, the same as the settings in the explanation generation task, the premise and the \emph{correct} hypothesis are taken as the input of GPT2$_{\text{CR-GE}}$ for generating explanations.
% As the model cannot produce a correct explanation with a wrong prediction, when evaluating the quality of generated explanations, 
% %we only consider the explanations with corresponding causal question correctly answered. In other words, 
% if a causal question is not correctly answered, the quality scores (e.g, BLEU, ROUGE, etc.) of the generated explanation are all set as 0. 

%if provide as correctness score the percentage of correct explanations in the subset of the first 100 examples where the predicted label was correct (80 in this experiment). We obtain a percentage of 34.68% correct explanations. While this percentage is low, we keep in mind that the selection criteria was only the accuracy of the label classifier and not the perplexity of the explanation. In the next experiment, we show how training (and selecting) only for generating explanations results in higher quality explanations.
%we denote as BART$_{\text{CR-GE}}$ and BART$_{\text{GE-CR}}$, respectively.

%\noindent \textbf{(a) Causal Reasoning Then Generating Explanation (CR-GE)}
%\noindent \textbf{(b) Generating Explanation Then Causal Reasoning (GE-CR)}

\noindent \textbf{Results} We measure the quality of generated explanations using the same automatic scores and human evaluation settings as the Explanation Generation experiment. The performance of causal reasoning is also measured using accuracy. 
The results are shown in Table~\ref{tab:res_mt}, where GPT2$_{\text{CR}}$ denotes the GPT2 model finetuned for the causal reasoning task, and GPT2$_{\text{EG}}$ refers to the GPT2 model finetuned for the explanation generation task. 
%We also list the results that the GPT2 model is provided with golden explanation (denoted as GPT2$_{\text{CR w/ Golden E}}$) as a celling of model performance. 
We observe that compared with GPT2$_{\text{CR}}$, the improved performance of GPT2$_{\text{CR-EG}}$ on causal reasoning indicates that the additional explanation can be helpful for the causal reasoning task, as it prompts model to have a deep understanding of the causal mechanisms. 
Interestingly, by comparing with GPT2$_{\text{EG}}$ and GPT2$_{\text{CR-EG}}$, we find that learning to predict the label can also be helpful for the explanation generation process. 
This indicates the synergistic effect of the causal reasoning and the explanation generation on promoting models' understanding of causal mechanism.

\subsection{Stability Analysis}

Previous studies indicate that models may utilize some superficial cues within the dataset to predict the label. This leads to the vulnerability of models when facing adversarial attacks \cite{poliak2018collecting,mccoy2019right}. Learning to generate the additional conceptual explanation may promote the understanding of causality to increase the stability of the reasoning model. Hence, we conduct a stability analysis to examine the specific effect of additional explanations.

Following \citet{bekoulis2018adversarial} and \citet{yasunaga2018robust}, we attack the causal reasoning system by adding a perturbation term on the word embeddings of inputs. The perturbation term is derived using the gradient-based FGM method \cite{miyato2016adversarial}. Table~\ref{tab:res_mt} shows the change of causal reasoning accuracy ($\Delta$Accu.) brought by the adversarial attack. For example, $\Delta = -6.40$ means a 6.40\% decrease of prediction accuracy after the adversarial attack. We find that, compared to the vanilla GPT2$_\text{CR}$ model, the explanation enhanced GPT2 model GPT2$_{\text{CR-EG}}$ demonstrates stronger stability. This suggests that, by training reasoning models to generate correct explanations of the causal facts, the understanding of the causality can be promoted, and then the stability of model performance can be increased.

%\subsection{Transfer Learning from e-CARE}
\subsection{Enhancing Pretrained Language Model with e-CARE}

Causal knowledge is critical for various NLP applications. In this section, we investigate if the causality knowledge provided by e-CARE can be used as a resource to boost model performance on other causal-related tasks.
%, other than just serving as a benchmark. 
To this end, we apply transfer learning by first finetuning a BERT model on e-CARE, then adapting the e-CARE-enhanced model (denoted as BERT$_\text{E}$) on a causal extraction task EventStoryLine 0.9 \cite{caselli2017event}, two causal reasoning tasks BECauSE 2.0 \cite{dunietz2017because} and COPA \cite{roemmele2011choice}, as well as a commonsense reasoning dataset CommonsenseQA \cite{talmor2019commonsenseqa}. On the EventStoryLine 0.9 dataset, we conduct experiment only on the instances about within-sentence causal relationship. The results are shown in Table~\ref{tab:res_trans}. We observe that the additional training process on e-CARE can consistently increase the model performance on all four tasks. This indicates the potential of e-CARE in providing necessary causality information for promoting causal-related tasks in multiple domains.

\begin{table}[t]
    \centering
    \small
    \begin{tabular}{lccc}
    \hline
    \textbf{Dataset} & \textbf{Metric} & \textbf{BERT}  & \textbf{BERT}$_{\text{E}}$\\ 
    \hline
    EventStoryLine 0.9 & F1 (\%) & 66.5 & \textbf{68.1} \\
    BECauSE 2.1 & Accu. (\%) & 76.8 & \textbf{81.0} \\
    COPA & Accu. (\%) & 70.4 & \textbf{75.4} \\
    CommonsenseQA & Accu. (\%) & 52.6 & \textbf{56.4} \\
    \hline
    %\textbf{Human Performance} & \\
    %\hline
    \end{tabular}
    \caption{Performance of e-CARE-enhanced BERT.}
    \label{tab:res_trans}
%\vspace{-0.1cm}
\end{table}

%Baselines We include baselines that rely on simple features to verify that ART is not trivially solvable due to noticeable annotation artifacts, observed in several crowdsourced datasets. The accuracies of all simple baselines are close to chance-performance on the task – indicating that the dataset is free of simple annotation artifacts.

\section{Discussion}

In this paper, we introduce additional explanation information for the causal reasoning process, and propose a corresponding explanation generation task. 
%Previous analysis further highlighted the interaction between the conceptual explanations
%In this paper, we propose an explanation generation task, which requires the model to generate an explanation for a certain causal fact that is as close as possible to the explanation of a given artificial annotation concept.
Previous literature concluded the explanation generation process as an \emph{abductive reasoning} process \cite{hanson1958patterns,peirce1974collected}  
%Formally, \emph{abductive reasoning} is defined as an inference to the most plausible explanation for the given observations. 
%\citet{peirce1974collected} further addressed the significance of abductive reasoning that ``abduction is the only logical operation which introduces any new ideas''. 
%Furthermore, they 
and highlighted the importance of the abdutive explanation generation, as it may interact with the causal reasoning process to promote the understanding of causal mechanism, and increase the efficiency and reliability of causal reasoning.

\begin{figure}
    \centering
    \includegraphics[width=0.99\linewidth]{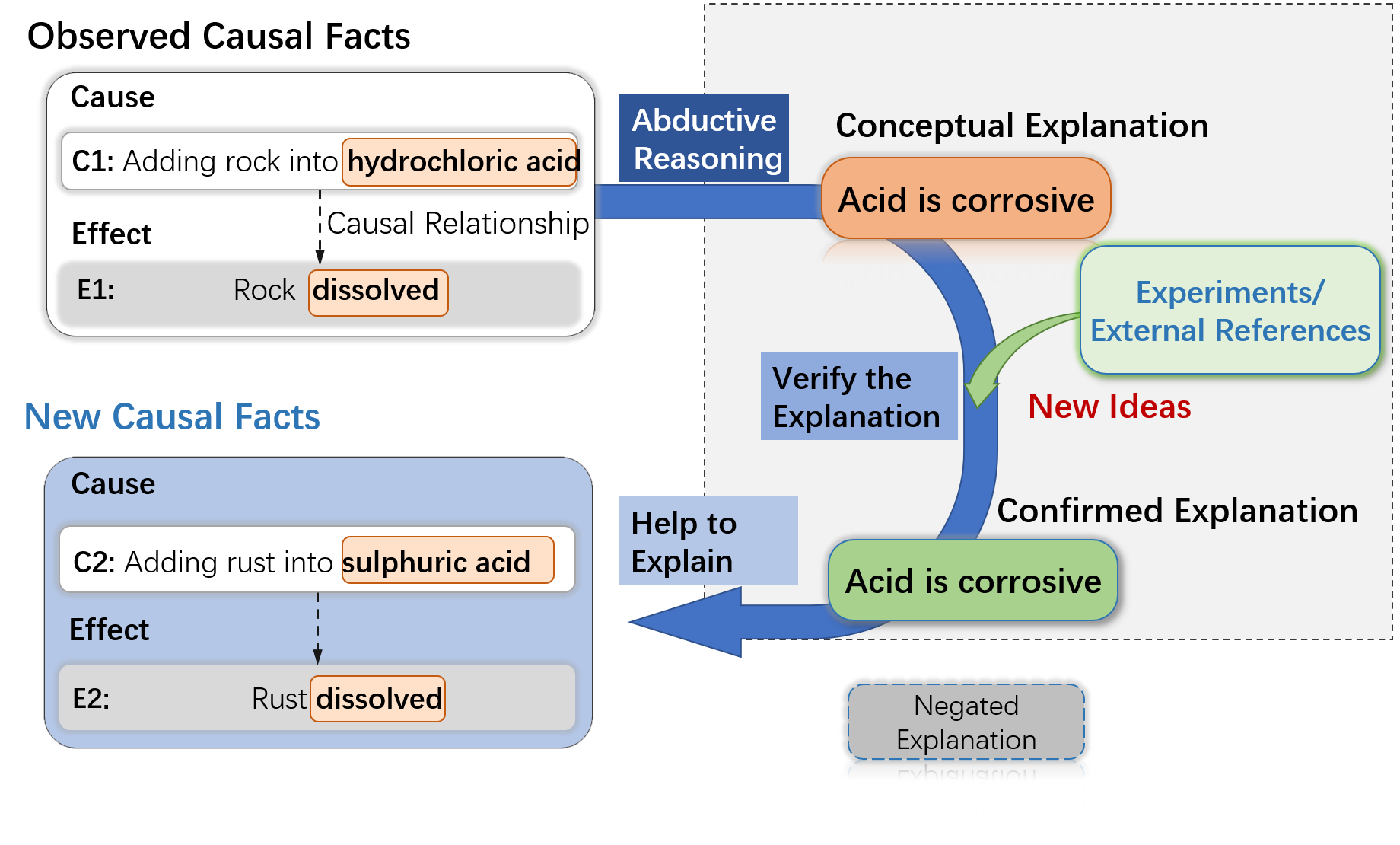}
    %\vspace{-0.8cm}
    %\setlength{\abovecaptionskip}{-0.1cm}
    %\setlength{\belowcaptionskip}{-0.2cm}
    \caption{Conceptual explanations of observed causality can be helpful for understanding the unseen causal facts.}
    \label{fig:discuss}
\end{figure}
%Therefore, by abdutively inferring the conceptual explanation, and employ the confirmed conceptual explanation to support the causal reasoning, 
%, and ``all the ideas of science come to it by way of abduction''. 

For example, as Figure~\ref{fig:discuss} shows, one may have an observation that $C_1$: \emph{adding rock into hydrochloric acid} caused $E_1$: \emph{rock dissolved}. Through abductive reasoning, one may come up with a conceptual explanation for the observation that \emph{acid is corrosive}. After that, one can confirm or rectify the explanation by experiments, or resorting to external references. In this way, new ideas about causality can be involved for understanding the observed causal fact. Then if the explanation is confirmed, it can be further utilized to support the causal reasoning process by helping to explain and validate other related causal facts, such as $C_2$: \emph{adding rust into sulphuric acid} may lead to $E_2$: {rust dissolved}.  
This analysis highlights 
%the complex cognitive and reasoning patterns of human beings in learning and inferring the causality. 
the pivotal role of conceptual explanation in learning and inferring causality. In this paper, we introduce the e-CARE dataset to provide causal explanations and support future research towards stronger human-like causal reasoning systems. 
%is a surmise to the abstract.
%that is crucial for AI systems.
%These arguments further highlights the in causal reasoning.

%We also argue that this abductive explanation generation process is critically different from the Abductive Natural Language Generation ($\alpha$NLG) task \cite{bhagavatula2019abductive}. In the $\alpha$NLG task, models are required to generate an event for a specific observed event, so that the generated event can explain why the specific observed event happens. Therefore, compared to the $\alpha$NLG task, our proposed abductive explanation generation task tests model's ability in obtaining deep and conceptual understandings of a set of observations based on its commonsense reasoning ability, rather than just fixing on explaining specific concrete events. 
%Which requires model to understanding the essential of causal fact and generate potential explanations. 
%In addition, we further propose an CEQ score, as an attempt on automatically evaluating the quality of abductively generated explanations.

\section{Conclusion}

In this paper, we present an \emph{e}xplainable CAusal REeasoning dataset \emph{e}-CARE, which contains over 21K causal questions, together with over 13K unique conceptual explanations about the deep understanding of the causal facts, which also makes e-CARE the largest causal reasoning benchmark. Experimental results show that both the causal reasoning task and especially the explanation generation task remain challenging for the SOTA pretrained language models. Moreover, the additional explanation signal can promote both the prediction accuracy and stability of models, highlighting the vital importance of the conceptual explanations in causal reasoning. 

\section{Acknowledgments}

We thank the anonymous reviewers for their constructive comments, and gratefully acknowledge the support of the New Generation Artificial Intelligence of China (2020AAA0106501), and the National Natural Science Foundation of China (62176079, 61976073).
%We hope that e-CARE can facilitate future researches towards more advanced causal reasoning systems with deeper understanding of the causal mechanism. 

%In this paper, we present the first study that investigating explainable causal reasoning. To support this task, we introduce an \emph{e}xplainable CAusal REeasoning dataset \emph{e}-CARE, which contains over 18K causal questions, together with over 10K explanations about the conceptual understanding of the causal facts, which also makes e-CARE temporarily the largest causal reasoning benchmark. In our experiments, we establish comprehensive baselines based on state-of-the-art pretrained language models, and show that both the causal reasoning and especially the explanation generation remains challenging for the SOTA pretrained language models. While the additional explanation signal can promote both the prediction accuracy and stability of model, highlighting the vital importance of conceptual explanations in causal reasoning. We hope that e-CARE will be valuable for future research on more advanced causal reasoning systems with stronger causality knowledge understanding and more complex causal reasoning capabilities. 

\bibliography{emnlp-ijcnlp-2021}

\begin{thebibliography}{50}
\expandafter\ifx\csname natexlab\endcsname\relax\def\natexlab#1{#1}\fi

\bibitem[{Bekoulis et~al.(2018)Bekoulis, Deleu, Demeester, and
  Develder}]{bekoulis2018adversarial}
Giannis Bekoulis, Johannes Deleu, Thomas Demeester, and Chris Develder. 2018.
\newblock Adversarial training for multi-context joint entity and relation
  extraction.
\newblock In \emph{Proceedings of the 2018 Conference on Empirical Methods in
  Natural Language Processing}, pages 2830--2836.

\bibitem[{Bethard and Martin(2008)}]{bethard2008learning}
Steven Bethard and James~H Martin. 2008.
\newblock Learning semantic links from a corpus of parallel temporal and causal
  relations.
\newblock In \emph{Proceedings of ACL-08: HLT, Short Papers}, pages 177--180.

\bibitem[{Bhagavatula et~al.(2019)Bhagavatula, Le~Bras, Malaviya, Sakaguchi,
  Holtzman, Rashkin, Downey, Yih, and Choi}]{bhagavatula2019abductive}
Chandra Bhagavatula, Ronan Le~Bras, Chaitanya Malaviya, Keisuke Sakaguchi, Ari
  Holtzman, Hannah Rashkin, Doug Downey, Wen-tau Yih, and Yejin Choi. 2019.
\newblock Abductive commonsense reasoning.
\newblock In \emph{International Conference on Learning Representations}.

\bibitem[{Bhakthavatsalam et~al.(2020)Bhakthavatsalam, Anastasiades, and
  Clark}]{bhakthavatsalam2020genericskb}
Sumithra Bhakthavatsalam, Chloe Anastasiades, and Peter Clark. 2020.
\newblock Genericskb: A knowledge base of generic statements.
\newblock \emph{arXiv preprint arXiv:2005.00660}.

\bibitem[{Camburu et~al.(2018)Camburu, Rockt{\"a}schel, Lukasiewicz, and
  Blunsom}]{camburu2018snli}
Oana-Maria Camburu, Tim Rockt{\"a}schel, Thomas Lukasiewicz, and Phil Blunsom.
  2018.
\newblock e-snli: Natural language inference with natural language
  explanations.
\newblock In \emph{NeurIPS}.

\bibitem[{Caselli and Vossen(2017)}]{caselli2017event}
Tommaso Caselli and Piek Vossen. 2017.
\newblock The event storyline corpus: A new benchmark for causal and temporal
  relation extraction.
\newblock In \emph{Proceedings of the Events and Stories in the News Workshop},
  pages 77--86.

\bibitem[{Chung et~al.(2014)Chung, Gulcehre, Cho, and
  Bengio}]{chung2014empirical}
Junyoung Chung, Caglar Gulcehre, KyungHyun Cho, and Yoshua Bengio. 2014.
\newblock Empirical evaluation of gated recurrent neural networks on sequence
  modeling.
\newblock \emph{arXiv preprint arXiv:1412.3555}.

\bibitem[{Cohen(1960)}]{cohen1960coefficient}
Jacob Cohen. 1960.
\newblock A coefficient of agreement for nominal scales.
\newblock \emph{Educational and psychological measurement}, 20(1):37--46.

\bibitem[{Devlin et~al.(2019)Devlin, Chang, Lee, and
  Toutanova}]{devlin2019bert}
Jacob Devlin, Ming-Wei Chang, Kenton Lee, and Kristina Toutanova. 2019.
\newblock Bert: Pre-training of deep bidirectional transformers for language
  understanding.
\newblock In \emph{Proceedings of the 2019 Conference of the North American
  Chapter of the Association for Computational Linguistics: Human Language
  Technologies, Volume 1 (Long and Short Papers)}, pages 4171--4186.

\bibitem[{DeYoung et~al.(2019)DeYoung, Jain, Rajani, Lehman, Xiong, Socher, and
  Wallace}]{deyoung2019eraser}
Jay DeYoung, Sarthak Jain, Nazneen~Fatema Rajani, Eric Lehman, Caiming Xiong,
  Richard Socher, and Byron~C Wallace. 2019.
\newblock Eraser: A benchmark to evaluate rationalized nlp models.
\newblock \emph{arXiv preprint arXiv:1911.03429}.

\bibitem[{Do et~al.(2011)Do, Chan, and Roth}]{do2011minimally}
Quang Do, Yee~Seng Chan, and Dan Roth. 2011.
\newblock Minimally supervised event causality identification.
\newblock In \emph{Proceedings of the 2011 Conference on Empirical Methods in
  Natural Language Processing}, pages 294--303.

\bibitem[{Dunietz et~al.(2017)Dunietz, Levin, and
  Carbonell}]{dunietz2017because}
Jesse Dunietz, Lori Levin, and Jaime~G Carbonell. 2017.
\newblock The because corpus 2.0: Annotating causality and overlapping
  relations.
\newblock In \emph{Proceedings of the 11th Linguistic Annotation Workshop},
  pages 95--104.

\bibitem[{Fellbaum(2010)}]{fellbaum2010wordnet}
Christiane Fellbaum. 2010.
\newblock Wordnet.
\newblock In \emph{Theory and applications of ontology: computer applications},
  pages 231--243. Springer.

\bibitem[{Girju et~al.(2007)Girju, Nakov, Nastase, Szpakowicz, Turney, and
  Yuret}]{girju2007semeval}
Roxana Girju, Preslav Nakov, Vivi Nastase, Stan Szpakowicz, Peter Turney, and
  Deniz Yuret. 2007.
\newblock Semeval-2007 task 04: Classification of semantic relations between
  nominals.
\newblock In \emph{Proceedings of the Fourth International Workshop on Semantic
  Evaluations (SemEval-2007)}, pages 13--18.

\bibitem[{Gururangan et~al.(2018)Gururangan, Swayamdipta, Levy, Schwartz,
  Bowman, and Smith}]{gururangan2018annotation}
Suchin Gururangan, Swabha Swayamdipta, Omer Levy, Roy Schwartz, Samuel Bowman,
  and Noah~A Smith. 2018.
\newblock Annotation artifacts in natural language inference data.
\newblock In \emph{Proceedings of the 2018 Conference of the North American
  Chapter of the Association for Computational Linguistics: Human Language
  Technologies, Volume 2 (Short Papers)}, pages 107--112.

\bibitem[{Hancock et~al.(2018)Hancock, Bringmann, Varma, Liang, Wang, and
  R{\'e}}]{hancock2018training}
Braden Hancock, Martin Bringmann, Paroma Varma, Percy Liang, Stephanie Wang,
  and Christopher R{\'e}. 2018.
\newblock Training classifiers with natural language explanations.
\newblock In \emph{Proceedings of the conference. Association for Computational
  Linguistics. Meeting}, volume 2018, page 1884. NIH Public Access.

\bibitem[{Hanson(1958)}]{hanson1958patterns}
Norwood~Russell Hanson. 1958.
\newblock \emph{Patterns of discovery: An inquiry into the conceptual
  foundations of science}, volume 251.
\newblock CUP Archive.

\bibitem[{Hendrickx et~al.(2019)Hendrickx, Kim, Kozareva, Nakov, S{\'e}aghdha,
  Pad{\'o}, Pennacchiotti, Romano, and Szpakowicz}]{hendrickx2019semeval}
Iris Hendrickx, Su~Nam Kim, Zornitsa Kozareva, Preslav Nakov, Diarmuid~O
  S{\'e}aghdha, Sebastian Pad{\'o}, Marco Pennacchiotti, Lorenza Romano, and
  Stan Szpakowicz. 2019.
\newblock Semeval-2010 task 8: Multi-way classification of semantic relations
  between pairs of nominals.
\newblock \emph{arXiv preprint arXiv:1911.10422}.

\bibitem[{Horgan(1995)}]{horgan1995complexity}
John Horgan. 1995.
\newblock From complexity to perplexity.
\newblock \emph{Scientific American}, 272(6):104--109.

\bibitem[{Jhamtani and Clark(2020)}]{jhamtani2020learning}
Harsh Jhamtani and Peter Clark. 2020.
\newblock Learning to explain: Datasets and models for identifying valid
  reasoning chains in multihop question-answering.
\newblock \emph{arXiv preprint arXiv:2010.03274}.

\bibitem[{Jonassen et~al.(2008)Jonassen, Ionas, and
  Ioan}]{jonassen2008designing}
David~H Jonassen, Ionas, and Gelu Ioan. 2008.
\newblock Designing effective supports for causal reasoning.
\newblock \emph{Educational Technology Research and Development},
  56(3):287--308.

\bibitem[{Lan et~al.(2019)Lan, Chen, Goodman, Gimpel, Sharma, and
  Soricut}]{lan2019albert}
Zhenzhong Lan, Mingda Chen, Sebastian Goodman, Kevin Gimpel, Piyush Sharma, and
  Radu Soricut. 2019.
\newblock Albert: A lite bert for self-supervised learning of language
  representations.
\newblock \emph{arXiv preprint arXiv:1909.11942}.

\bibitem[{Lewis et~al.(2020)Lewis, Liu, Goyal, Ghazvininejad, Mohamed, Levy,
  Stoyanov, and Zettlemoyer}]{lewis2020bart}
Mike Lewis, Yinhan Liu, Naman Goyal, Marjan Ghazvininejad, Abdelrahman Mohamed,
  Omer Levy, Veselin Stoyanov, and Luke Zettlemoyer. 2020.
\newblock Bart: Denoising sequence-to-sequence pre-training for natural
  language generation, translation, and comprehension.
\newblock In \emph{Proceedings of the 58th Annual Meeting of the Association
  for Computational Linguistics}, pages 7871--7880.

\bibitem[{Li et~al.(2020)Li, Ding, Liu, Hu, and Van~Durme}]{li2020guided}
Zhongyang Li, Xiao Ding, Ting Liu, J~Edward Hu, and Benjamin Van~Durme. 2020.
\newblock Guided generation of cause and effect.
\newblock IJCAI.

\bibitem[{Lin(2004)}]{lin2004rouge}
Chin-Yew Lin. 2004.
\newblock Rouge: A package for automatic evaluation of summaries.
\newblock In \emph{Text summarization branches out}, pages 74--81.

\bibitem[{Liu et~al.(2019)Liu, Ott, Goyal, Du, Joshi, Chen, Levy, Lewis,
  Zettlemoyer, and Stoyanov}]{liu2019roberta}
Yinhan Liu, Myle Ott, Naman Goyal, Jingfei Du, Mandar Joshi, Danqi Chen, Omer
  Levy, Mike Lewis, Luke Zettlemoyer, and Veselin Stoyanov. 2019.
\newblock Roberta: A robustly optimized bert pretraining approach.

\bibitem[{Luo et~al.(2016)Luo, Sha, Zhu, Hwang, and Wang}]{luo2016commonsense}
Zhiyi Luo, Yuchen Sha, Kenny~Q Zhu, Seung-won Hwang, and Zhongyuan Wang. 2016.
\newblock Commonsense causal reasoning between short texts.
\newblock In \emph{KR}, pages 421--431.

\bibitem[{McCoy et~al.(2019)McCoy, Pavlick, and Linzen}]{mccoy2019right}
Tom McCoy, Ellie Pavlick, and Tal Linzen. 2019.
\newblock Right for the wrong reasons: Diagnosing syntactic heuristics in
  natural language inference.
\newblock In \emph{Proceedings of the 57th Annual Meeting of the Association
  for Computational Linguistics}, pages 3428--3448.

\bibitem[{Mirza et~al.(2014)Mirza, Sprugnoli, Tonelli, and
  Speranza}]{mirza2014annotating}
Paramita Mirza, Rachele Sprugnoli, Sara Tonelli, and Manuela Speranza. 2014.
\newblock Annotating causality in the tempeval-3 corpus.
\newblock In \emph{EACL 2014 Workshop on Computational Approaches to Causality
  in Language (CAtoCL)}, pages 10--19. Association for Computational
  Linguistics.

\bibitem[{Miyato et~al.(2016)Miyato, Dai, and
  Goodfellow}]{miyato2016adversarial}
Takeru Miyato, Andrew~M Dai, and Ian Goodfellow. 2016.
\newblock Adversarial training methods for semi-supervised text classification.
\newblock \emph{arXiv preprint arXiv:1605.07725}.

\bibitem[{Mostafazadeh et~al.(2016)Mostafazadeh, Grealish, Chambers, Allen, and
  Vanderwende}]{mostafazadeh2016caters}
Nasrin Mostafazadeh, Alyson Grealish, Nathanael Chambers, James Allen, and Lucy
  Vanderwende. 2016.
\newblock Caters: Causal and temporal relation scheme for semantic annotation
  of event structures.
\newblock In \emph{Proceedings of the Fourth Workshop on Events}, pages 51--61.

\bibitem[{Ning et~al.(2019)Ning, Feng, Wu, and Roth}]{ning2019joint}
Qiang Ning, Zhili Feng, Hao Wu, and Dan Roth. 2019.
\newblock Joint reasoning for temporal and causal relations.
\newblock \emph{arXiv preprint arXiv:1906.04941}.

\bibitem[{Papineni et~al.(2002)Papineni, Roukos, Ward, and
  Zhu}]{papineni2002bleu}
Kishore Papineni, Salim Roukos, Todd Ward, and Wei-Jing Zhu. 2002.
\newblock Bleu: a method for automatic evaluation of machine translation.
\newblock In \emph{Proceedings of the 40th annual meeting of the Association
  for Computational Linguistics}, pages 311--318.

\bibitem[{Peirce(1974)}]{peirce1974collected}
Charles~Sanders Peirce. 1974.
\newblock \emph{Collected papers of charles sanders peirce}, volume~2.
\newblock Harvard University Press.

\bibitem[{Perez et~al.(2019)Perez, Karamcheti, Fergus, Weston, Kiela, and
  Cho}]{perez2019finding}
Ethan Perez, Siddharth Karamcheti, Rob Fergus, Jason Weston, Douwe Kiela, and
  Kyunghyun Cho. 2019.
\newblock Finding generalizable evidence by learning to convince q\&a models.
\newblock \emph{arXiv preprint arXiv:1909.05863}.

\bibitem[{Poliak et~al.(2018)Poliak, Haldar, Rudinger, Hu, Pavlick, White, and
  Van~Durme}]{poliak2018collecting}
Adam Poliak, Aparajita Haldar, Rachel Rudinger, J~Edward Hu, Ellie Pavlick,
  Aaron~Steven White, and Benjamin Van~Durme. 2018.
\newblock Collecting diverse natural language inference problems for sentence
  representation evaluation.
\newblock In \emph{Proceedings of the 2018 Conference on Empirical Methods in
  Natural Language Processing}, pages 67--81.

\bibitem[{Radford et~al.(2018)Radford, Narasimhan, Salimans, and
  Sutskever}]{radford2018improving}
Alec Radford, Karthik Narasimhan, Tim Salimans, and Ilya Sutskever. 2018.
\newblock Improving language understanding by generative pre-training.

\bibitem[{Radford et~al.(2019)Radford, Wu, Child, Luan, Amodei, and
  Sutskever}]{radford2019language}
Alec Radford, Jeffrey Wu, Rewon Child, David Luan, Dario Amodei, and Ilya
  Sutskever. 2019.
\newblock Language models are unsupervised multitask learners.
\newblock \emph{OpenAI blog}, 1(8):9.

\bibitem[{Rajani et~al.(2019)Rajani, McCann, Xiong, and
  Socher}]{rajani2019explain}
Nazneen~Fatema Rajani, Bryan McCann, Caiming Xiong, and Richard Socher. 2019.
\newblock Explain yourself! leveraging language models for commonsense
  reasoning.
\newblock In \emph{Proceedings of the 57th Annual Meeting of the Association
  for Computational Linguistics}, pages 4932--4942.

\bibitem[{Roemmele et~al.(2011)Roemmele, Bejan, and
  Gordon}]{roemmele2011choice}
Melissa Roemmele, Cosmin~Adrian Bejan, and Andrew~S Gordon. 2011.
\newblock Choice of plausible alternatives: An evaluation of commonsense causal
  reasoning.
\newblock In \emph{AAAI Spring Symposium: Logical Formalizations of Commonsense
  Reasoning}, pages 90--95.

\bibitem[{Sakaguchi et~al.(2020)Sakaguchi, Le~Bras, Bhagavatula, and
  Choi}]{sakaguchi2020winogrande}
Keisuke Sakaguchi, Ronan Le~Bras, Chandra Bhagavatula, and Yejin Choi. 2020.
\newblock Winogrande: An adversarial winograd schema challenge at scale.
\newblock In \emph{Proceedings of the AAAI Conference on Artificial
  Intelligence}, volume~34, pages 8732--8740.

\bibitem[{Sap et~al.(2019)Sap, Le~Bras, Allaway, Bhagavatula, Lourie, Rashkin,
  Roof, Smith, and Choi}]{sap2019atomic}
Maarten Sap, Ronan Le~Bras, Emily Allaway, Chandra Bhagavatula, Nicholas
  Lourie, Hannah Rashkin, Brendan Roof, Noah~A Smith, and Yejin Choi. 2019.
\newblock Atomic: An atlas of machine commonsense for if-then reasoning.
\newblock In \emph{Proceedings of the AAAI Conference on Artificial
  Intelligence}, volume~33, pages 3027--3035.

\bibitem[{Sembugamoorthy and
  Chandrasekaran(1986)}]{sembugamoorthy1986functional}
V~Sembugamoorthy and B~Chandrasekaran. 1986.
\newblock Functional representation of devices and compilation of diagnostic
  problem-solving systems.
\newblock \emph{Experience, memory and Reasoning}, pages 47--73.

\bibitem[{Speer and Havasi(2013)}]{speer2013conceptnet}
Robert Speer and Catherine Havasi. 2013.
\newblock Conceptnet 5: A large semantic network for relational knowledge.
\newblock In \emph{The People’s Web Meets NLP}, pages 161--176. Springer.

\bibitem[{Talmor et~al.(2019)Talmor, Herzig, Lourie, and
  Berant}]{talmor2019commonsenseqa}
Alon Talmor, Jonathan Herzig, Nicholas Lourie, and Jonathan Berant. 2019.
\newblock Commonsenseqa: A question answering challenge targeting commonsense
  knowledge.
\newblock In \emph{Proceedings of the 2019 Conference of the North American
  Chapter of the Association for Computational Linguistics: Human Language
  Technologies, Volume 1 (Long and Short Papers)}, pages 4149--4158.

\bibitem[{Waldmann and Hagmayer(2013)}]{waldmann2013causal}
Michael~R Waldmann and York Hagmayer. 2013.
\newblock Causal reasoning.

\bibitem[{Wiegreffe and Marasovi{\'c}(2021)}]{wiegreffe2021teach}
Sarah Wiegreffe and Ana Marasovi{\'c}. 2021.
\newblock Teach me to explain: A review of datasets for explainable nlp.
\newblock \emph{arXiv preprint arXiv:2102.12060}.

\bibitem[{Yang et~al.(2019)Yang, Dai, Yang, Carbonell, Salakhutdinov, and
  Le}]{yang2019xlnet}
Zhilin Yang, Zihang Dai, Yiming Yang, Jaime Carbonell, Ruslan Salakhutdinov,
  and Quoc~V Le. 2019.
\newblock Xlnet: Generalized autoregressive pretraining for language
  understanding.
\newblock \emph{arXiv preprint arXiv:1906.08237}.

\bibitem[{Yasunaga et~al.(2018)Yasunaga, Kasai, and Radev}]{yasunaga2018robust}
Michihiro Yasunaga, Jungo Kasai, and Dragomir Radev. 2018.
\newblock Robust multilingual part-of-speech tagging via adversarial training.
\newblock In \emph{Proceedings of the 2018 Conference of the North American
  Chapter of the Association for Computational Linguistics: Human Language
  Technologies, Volume 1 (Long Papers)}, pages 976--986.

\bibitem[{Ye et~al.(2020)Ye, Huang, and Ren}]{ye2020teaching}
Qinyuan Ye, Xiao Huang, and Xiang Ren. 2020.
\newblock Teaching machine comprehension with compositional explanations.
\newblock \emph{arXiv preprint arXiv:2005.00806}.

\end{thebibliography}
\bibliographystyle{acl_natbib}

\section{More Discussions about the e-CARE Dataset}

\subsection{The Generality of the Conceptual Explanation}

\begin{table*}[t]
    \centering
    \scriptsize
    \begin{tabular}{l|l}
    \toprule
    \textbf{(a)} \textit{Premise}: Tom held a copper block by hand and heated it on fire. & \textbf{(b)} \textit{Premise}:This computer's heat dispersion performance is bad. \\
    \quad \ \ \textit{Ask-for}: Effect & \quad \ \ \textit{Ask-for}: Effect \\
    \quad \ \ \textit{Hypothesis 1}: His fingers felt burnt for a short time. (\Checkmark) & \quad \ \ \textit{Hypothesis 1}: Designers add copper tubes into the computer. (\Checkmark) \\
    \quad \ \ \textit{Hypothesis 2}: The copper block kept the same. (${\times}$) & \quad \ \ \textit{Hypothesis 2}: Designers put the computer into the ice water. (${\times}$) \\
    \hline
    \quad \ \ \textit{Explanation}: \textbf{Copper is a good thermal conductor.} & \quad \ \ \textit{Explanation}: \textbf{Copper is a good thermal conductor.} \\ 
    \bottomrule
    \end{tabular}
    \caption{Two instances from the e-CARE dataset. }
    \label{tab:exp_two}
%\vspace{-0.1cm}
\end{table*}

In this paper, we construct the dataset by first obtaining the conceptual explanations, then obtaining the causal questions. This is because, we also hope to find the conceptual explanations with more generality, that that can explain more than one causal fact, but can explain a set of correlated causal facts. Table~\ref{tab:exp_two} demonstrate an example of such conceptual explanation. The \emph{explanation} points out the nature of Copper that \emph{Copper is a good thermal conductor}, so that holding copper on fire will make fingers feel burnt immediately. Additionally, the same explanation can also provide insights about another causal fact seemingly totally different from the case in Table~\ref{tab:exp}~(a), that putting copper tubes into computer can promote thermal dispersion. This is because, the conceptual explanation points out the nature of copper, which drives a set of causal facts into existence.

This example demonstrate the usefulness of the conceptual explanations in providing the deep understanding of causality to support the causal reasoning. However, note that in this paper, we do not assume all the statements we collected can explain multiple causal facts. Instead, we resort to the empirical knowledge of human annotators to find such explanations. Specifically, we distribute statements to several annotators, and require each annotator to generate a causal fact that can be explained by the statement. For a certain statement, if it is distributed to multiple annotators and more than one annotator can generate a corresponding causal fact, then we assume that this statement can be a conceptual statement.   

\subsection{The Exhaustiveness of the Explanations}

Another point we wish to elucidate is about the exhaustiveness of the explanations. In this paper, we only aim at providing \emph{plausible} explanations that can explain the causal fact, but do not assume the provided explanations to be exhaustive or self-sufficient.  

\subsection{The Relationship between the Unique Explanations and Causal Questions} Due to the practical limits, to ensure the coverage of dataset, only a part of statements are distributed to multiple annotators, as described in Section 3.1.  

\section{Data Collection Details}

\subsection{Collection of Explanations} We collect the potential explanations from a commonsense knowledge base GenericsKB \cite{bhakthavatsalam2020genericskb}, which contains naturally occurring generic statements, such as ``Trees remove carbon dioxide from the atmosphere'', collected from multiple corpora. We first filtered the statements according to their quality score $s$, which is a human-annotation based metric, provided in the GenericsKB and evaluating the correctness of each statement. To ensure the factual correctness of the potential explanations, we only kept the statements whose quality score are among the highest 1\%. In addition, we also excluded the statements including: (1) Overly complex statements. The statements with connective, and statements with more than 20 words are excluded. This is because, by observation, we found that the annotators always struggle with understand and generate plausible causal facts for the over complex explanations. The number 20 is an empirical setting. (2) Statements describing named entities.  (3) Statements describing the hypernymy or hyperonymy relationship between the subject and object. For example, the statement \emph{Monkey is a kind of mammal. } describes the hypernymy relationship between the subject monkey and object mammal. This kind of statement does not belong to the three kinds of information that a valid explanation contains, as mentioned in Section 3.1.

After the filtering process, totally 19K statements are remained to be the potential explanations. Note that we do not assume that the statements after the filtering process are necessarily to be valid potential explanation and force the annotators to generate corresponding causal fact(s). Instead, we left the judgment to the annotators. If a statement has already been distributed to three annotators and no annotator can generate a corresponding causal question for this statement, then it is discarded. 
%\iffalse
\subsection{Collection of Causal Questions} 

We guided the annotators using illustrative examples to avoid the following mistakes:

\noindent (1) The generated cause and effect cannot be explained by the statement.

\begin{itemize}
 \item Wrong Case 

\emph{Explanation}: Copper is a good The copper block was oxidized and the surface became dark.. \\
\emph{Cause}: Tom held a copper block and heated it on fire.  \\
\emph{Effect}: The copper block was oxidized and the surface became dark.  \\

 \item Correct Case 

\emph{Explanation}: Copper is a good thermal conductor. \\
\emph{Cause}: Tom held a copper block by hand and heated it on fire.  \\
\emph{Effect}: His fingers felt burnt for a short time. \\
\end{itemize}

\noindent (2) The generated ``cause'' and ``effect'' do not form a valid causal relationship.

\begin{itemize}
 \item Wrong Case 

\emph{Explanation}: Oncologists specialize in the treatment of cancer. \\
\emph{Cause}: Jerry suffered from cancer. \\
\emph{Effect}: Jerry consulted many artists. \\

\item Correct Case 
 
\emph{Explanation}: Oncologists specialize in the treatment of cancer. \\
\emph{Cause}: Jerry suffered from cancer. \\
\emph{Effect}: Jerry consulted many oncologists. \\

\end{itemize}

\noindent (3) The distractor can also form a causal relationship with the premise.

\begin{itemize}
\item Wrong Case 

\emph{Explanation}: Oncologists specialize in the treatment of cancer. \\
\emph{Cause}: Jerry suffered from cancer. \\
\emph{Effect}: Jerry consulted many oncologists. \\
\emph{Disctractor Cause}: Jerry consulted many traditional herbalists. \\
\end{itemize}

\noindent (4) The generated distractor is uninformative. 
\begin{itemize}
\item Wrong Case

\emph{Explanation}: Copper is a good thermal conductor. \\
\emph{Cause}: Tom held a copper block by hand and heated it on fire.  \\
\emph{Effect}: His fingers felt burnt for a short time. \\
\emph{Disctractor Effect}: His fingers did not feel burnt for a short time.
\end{itemize}

%\fi
\section{Adversarial Filtering} 

During the annotation process, some superficial clues may be incurred into the dataset, which makes the correct and implausible hypothesis can be distinguished merely using these annotation artifacts. To decrease the influence of potential annotation artifacts, we introduce an Adversarial Filtering algorithm \cite{bhagavatula2019abductive} to refine our dataset. 

In specific, for an arbitrary causal question $\langle p, a, h^+, h^- \rangle$, where $p$ is the premise, $a \in [``cause'', ``effect'']$ is an ask-for annotator, $h^+$ and $h^-$ is the correct and wrong hypothesis, respectively, if $\langle p, h^+\rangle$ and $\langle p, h^- \rangle$ can be easily distinguished by a predictive model, then we replace $h^-$ with another implausible hypothesis $h^{-'}$ sampled from an implausible hypothesis set $\mathcal{H}$, so that $\langle p, h^{-'}\rangle$ is harder to be distinguished from $\langle p, h^+ \rangle$. Where the implausible hypothesis set $\mathcal{H}$ is the collection of all wrong hypotheses within the dataset. 

Algorithm 1 provides a formal description of our adversarial filtering algorithm. Specifically, in each iteration $i$, we randomly split the dataset into a training set $\mathcal{T}_i$ and a validation set $\mathcal{V}_i$. Then a model $\mathcal{M}_i$ is trained on $\mathcal{T}_i$ to update $\mathcal{V}_i$ to make it more challenging for $\mathcal{M}_i$. To this end, given an instance $\langle p_j, a_j, h^+_j, h^-_{j0} \rangle \in \mathcal{V}_i$, we randomly sample $K$ more implausible hypotheses $h^-_j1',\cdots, h^-_jK'$. Let $\delta^{\mathcal{M}_i}_k$ denotes the difference of model evaluation between $\langle p_j, a_j, h^+_j, h^-_j \rangle$ and $\langle p_j, a_j, h^-_k \rangle$, where $\delta^{\mathcal{M}_i}_k<0$ means model $\mathcal{M}_i$ favors $h^+_j$ to be the plausible hypothesis than the implausible hypothesis $h^-_{jk}$. With probability $t_i$, we replace $h^-_j$ with the implausible that is hardest to distinguish with $h^+_j$, i.e., $h^-_j=h^-_{jl}$, $l=\mathop{\arg\min}_{l}\delta^{\mathcal{M}_i}_k$.  
In this way, in each iteration, the proportion of easy implausible hypotheses decreases, and then the adversary model is forced to capture more causality knowledge.

\begin{algorithm}
\small
\caption{Adversarial Filtering}
\begin{algorithmic}[1] %每行显示行号
    \Require number of iteration $n$, dataset $\mathcal{D}_0$, implausible hypothesis set $\mathcal{H}^-$, initial and final temperature parameter $t_s$ and $t_e$.
    \Ensure dataset $\mathcal{D}_n$
    \For{iteration $i=1 \to (n-1)$}
        \State $t_i=t+e+\frac{t_s-t_e}{1+e^{0.3(i-3n/4)}}$
        \State Random split $M_i$ into training set $\mathcal{T}_i$ and validation set $\mathcal{V}_i$
        \State Train Model $M_i$ on $\mathcal{T}_i$
        \For{instance $j \in \mathcal{S}_i$ }
            \For{$h^-_{jk}\in \mathcal{H}^-_j$}
                \State Calculate $\delta^{\mathcal{M}_i}_k(\langle p_j, a_j, h^+_j \rangle, \langle p_j, a_j, h^-_{jk} \rangle)$
            \State $l=\mathop{\arg\min}_{l}\delta^{\mathcal{M}_i}_k$
            \State Sample $r$ from a Uniform distribution $U(0,1)$
            \State If $r<t_i$ or $\delta^{\mathcal{M}_i}_l<0$ then $h^-_j=h^-_{jl}$ 
            \State Add instance $j$ into $\mathcal{S}_i$
            \EndFor
        \EndFor
    \EndFor
    \State $\mathcal{D}_n=\mathcal{S}_n$
\end{algorithmic}
\end{algorithm}

We implemented the adversary model using pretrained language model RoBERTa-base \cite{liu2019roberta}.
The AF algorithm is run for 25 iterations and the temperature $t_i$ follows a sigmoid function, parameterized by the iteration number, between $t_s = 1.0$ and $t_e = 0.2$. For each instance, we sampled $K=20$ more implausible hypotheses from the implausible hypothesis set $\mathcal{H}$.

\begin{table*}
    \centering
    \small
    \begin{tabular}{cc}
    \hline
    \textbf{Model} \textbf{Input Format} \\
    \hline
    GPT2 & $<|$startoftext$|>$ C [SEP] E $<|$endoftext$|>$ \\
    RoBERTa & $<$s$>$ C $<$\/s$>$ E $<$\/s$>$  \\
    BART & $<$s$>$ C $<$\/s$>$ E $<$\/s$>$ \\
    XLNET & $<$cls$>$ C $<$sep$>$ E $<$sep$>$  \\
    BERT & [CLS] C [SEP] E [SEP] \\
    ALBERT & [CLS] C [SEP] E [SEP]  \\
    \hline
    \end{tabular}
    \caption{Input format of models in the causal reasoning task.}
    \label{tab:input_format}
\end{table*}

\section{Details of Experiments}
\subsection{Details of the Causal Reasoning Experiment}
\noindent \textbf{Settings} In this paper, the causal reasoning task is defined as a multiple-choice problem, which requires the model to choose a more plausible hypothesis from two candidates, so that the premise and hypothesis can form a valid causal fact. 
Therefore, the causal reasoning task could be formalized as a prediction problem: given a candidate cause fact $\langle cause, effect \rangle$ composed of the premise event and one of the hypothesis events, the prediction model is required to predict a score measuring the causality of the event pair. Note that the ask-for indicator decides whether the premise or candidate hypothesis to be the cause or effect, respectively.

To this end, we concatenate the premise with each one of the candidate hypothesis to form two candidate causal facts. Then each of the candidate causal fact is fed into the models, to obtain a probability measuring the plausibility of the candidate causal fact. To satisfy the input format of the pretrained language models, the input candidate causal fact is preprocessed by adding special tokens. Additionally, we adapt GPT2 and BART to predictive causal reasoning task by adding an $\text{EOS}$ token to the end of input text, and making predictions based on the representation of the $\text{EOS}$ token. The specific input format of the models is listed in Table~\ref{tab:input_format}, where $C$, $E$ denotes the cause and effect of the candidate causal fact, respectively.

%To this end, we concatenate the premise with each one of the candidate hypothesis to form two candidate causal facts. Then each of the candidate causal fact is fed into the models, to obtain a probability measuring the plausibility of the candidate causal fact. To satisfy the input format of the pretrained language models, the input candidate causal fact is preprocessed by adding special tokens. Additionally, we adapt GPT2 and BART to predictive causal reasoning task by adding an $\text{EOS}$ token to the end of input text, and making predictions based on the representation of the $\text{EOS}$ token. The specific input format of the models is listed in Table~\ref{tab:input_format}, where $C$, $E$ denotes the cause and effect of the candidate causal fact, respectively.

\noindent \textbf{Training Details} In the causal reasoning task, we optimize all the models with a batch size of 64, learning rate of 1e-5, and the model is finetuned for 3 epochs. 

\subsection{Details of the Explanation Generation Experiment}
\noindent \textbf{Settings} In the explanation generation experiment, models are trained to generate an explanation for a given valid causal fact $\langle C, E\rangle$. Hence, the input of GPT2 is formated as:
\begin{equation}
\small
<|startoftext|> C \ [SEP] \ E <|endoftext|>, 
\end{equation}
\noindent where  $<|$startoftext$|>$ and $<|$endoftext$|>$ are two special tokens. The input of the GRU-Seq2Seq model is formated as: 
\begin{equation}
\small
<SOS> C \ , \ E <EOS>.
\end{equation}
\noindent \textbf{Training Details} 
In the explanation generation task, the GPT2 model is trained with a batch size of 32, learning rate of 1e-5, and the model is finetuned for 10 epochs. For the GRU-Seq2seq model, both the encoder and the decoder contains 2 GRU layers with a dimension of 300$\times$300. The word embedding is initialized using 300-dimension GloVe. During optimazation, the GRU-Seq2seq model is trained for 10 epochs as well.

\subsection{Details of Explanation AND Generation Experiment}
\noindent \textbf{Settings} 
Given a causal question, we first concatenate the premise with each one of the candidate hypothesis to form two candidate causal facts. Then each of the candidate causal fact is fed into the GPT2 model, to get a distributed representation of the candidate causal fact. Then probability measuring the plausibility of the candidate causal fact is predicted using an MLP based on the distributed representation. After predicting plausibility score of two candidate causal facts, the model is trained to generate an explanation based on only the representation of the candidate causal fact that model thinks is more likely to be valid. 

\noindent \textbf{Training Details} 
During the training process, to balance the generation loss and prediction loss, we introduce an balance coefficient $\lambda$. Hence, the loss function is formulated as $L=(1-\lambda)L_{\text{Prediction}}+\lambda L_{\text{Generation}}$. We empirically set $\lambda=0.1$. The batch size and learning rate are also set as 32 and 1e-5, respectively. While different to the explanation generation process, in the Generate And Prediction experiment, the GPT2 model is trained for 5 epochs, as it receives two kinds of supervision signals.

\subsection{Details of Transfer Analysis}
\noindent \textbf{Settings} 

All four tasks in the transfer analysis can be formalized as multiple-choice problem. Specifically, the causal event extraction task EventStoryLine requires model to predict whether two phrase-level events within a sentence can form a causal relationship. While in two causal reasoning tasks BECauSE 2.0 \cite{dunietz2017because} and COPA \cite{roemmele2011choice}, models are required to choose a plausible hypothesis, so that the premise and the hypothesis can form a valid causal fact. The CommonsenseQA \cite{talmor2019commonsenseqa} task requires model to choose a correct answer for a given question. We list the specific format of the input on these four tasks in Table~\ref{tab:input_tf}, where $C$ and $E$ denotes the cause and effect, respectively, $Q$ and $A$ denotes the question and answer, respectively.  

\begin{table}[t]
    \centering
    \small
    \begin{tabular}{cc}
    \hline
    \textbf{Dataset} \textbf{Input Format} \\
    \hline
    EventStoryLine & [CLS] Statement \\
    BECauSE 2.0 & [CLS] C [SEP] E [SEP] \\
    COPA & [CLS] C [SEP] E [SEP]  \\
    CommonsenseQA 2.0 & [CLS] Q [SEP] A [SEP] \\
    \hline
    \end{tabular}
    \caption{Input format of models in the transfer analysis.}
    \label{tab:input_tf}
%\vspace{-0.1cm}
\end{table}

\noindent \textbf{Training Details} 
To equip model with the causality knowledge within e-CARE, we train a BERT model for 3 epochs, with a batch size of 32 and a learning rate of 1e-5. Then in the following finetuning stage, on all four datasets, both BERT and e-CARE enhanced model BERT$_{\text{E}}$ are fine-tuned using a grid search with the following set of hyper-parameters:

\begin{itemize}
\item batch size: \{16, 32\}
\item number of epochs: \{3,5,10\}
\item learning rate: \{1e-6, 1e-5\}
\end{itemize}

\end{document}